\pdfoutput=1
\documentclass[11pt]{article}

\usepackage{emnlp2021}

\usepackage{times}
\usepackage{latexsym}
\usepackage[T1]{fontenc}
\usepackage[utf8]{inputenc}

\usepackage{microtype}

\usepackage{times}
\usepackage{multirow}
\usepackage{latexsym}
\usepackage{graphicx}
\usepackage{amsmath}
\usepackage{booktabs, array}
\usepackage{tabularx, ragged2e}
\usepackage[inline]{enumitem}
\usepackage{lipsum,xcolor}
\usepackage{todonotes}
\usepackage{hyperref}
\usepackage{xspace}
\usepackage{tablefootnote}
\usepackage{paralist}
\newcommand{\Fbox}[1]{\fbox{\strut#1}}
\newcommand\dunderline[3][-1pt]{{%
  \sbox0{#3}%
  \ooalign{\copy0\cr\rule[\dimexpr#1-#2\relax]{\wd0}{#2}}}}

\newcommand{\gluecos}{{\textsc{Gluec\scriptsize{o}}\textsc{s}}\xspace}
\newcommand{\eng}{{\textsc{En Sing-Task}}\xspace}
\newcommand{\hindi}{{\textsc{Hi Sing-Task}}\xspace}
\newcommand{\xsing}{{\textsc{X Sing-Task}}\xspace}
\newcommand{\bilingual}{{\textsc{Hi-En Sing-Task}}\xspace}
\newcommand{\xbilingual}{{\textsc{X-En Sing-Task}}\xspace}
\newcommand{\dualbilingual}{{\textsc{Hi-En/Nli-Qa Multi-Task}}\xspace}
\newcommand{\mlmbilingual}{{\textsc{Hi-En/Mlm Multi-Task}}\xspace}
\newcommand{\generalcs}{\textsc{Gen-cs}\xspace}
\newcommand{\moviecs}{\textsc{Movie-cs}\xspace}
\newcolumntype{C}{>{\arraybackslash}X}

\newcommand{\HiEn}{\textsc{Hi-En}\xspace}
\newcommand{\EsEn}{\textsc{Es-En}\xspace}
\newcommand{\TaEn}{\textsc{Ta-En}\xspace}
\newcommand{\MlEn}{\textsc{Ml-En}\xspace}

\newcommand{\marked}[1]{\textcolor{blue}{\textit{#1}}}
\newcommand{\addmarked}[1]{\textcolor{purple}{\textit{#1}}}

\newcommand\footnoteref[1]{\protected@xdef\@thefnmark{\ref{#1}}\@footnotemark}
\title{The Effectiveness of Intermediate-Task Training for  \\ Code-Switched Natural Language Understanding}

\author{Archiki Prasad, Mohammad Ali Rehan, Shreya Pathak, Preethi Jyothi \\
         Indian Institute of Technology, Bombay}
\begin{document}
\maketitle
\begin{abstract}
%Code-switching is a widely prevalent linguistic phenomenon in multilingual societies. 
While recent benchmarks have spurred a lot of new work on improving the generalization of pretrained multilingual language models on multilingual tasks, techniques to improve code-switched natural language understanding tasks have been far less explored. In this work, we propose the use of \emph{bilingual intermediate pretraining} as a reliable technique to derive large and consistent performance gains on three different NLP tasks using code-switched text. We achieve substantial absolute improvements of $7.87\%$, $20.15\%$, and $10.99\%$, on the mean accuracies and F1 scores over previous state-of-the-art systems for Hindi-English Natural Language Inference (NLI), Question Answering (QA) tasks, and Spanish-English Sentiment Analysis (SA) respectively. We show consistent performance gains on four different code-switched language-pairs (Hindi-English , Spanish-English, Tamil-English and Malayalam-English) for SA. We also present a code-switched masked language modeling (MLM) pretraining technique that consistently benefits SA compared to standard MLM pretraining using real code-switched text.
%We also provide detailed analyses of the impact of translation and transliteration quality, and the effectiveness of MLM pretraining using real code-switched text.
% This document is a supplement to the general instructions for *ACL authors. It contains instructions for using the \LaTeX{} style files for EMNLP 2021. 
% The document itself conforms to its own specifications, and is therefore an example of what your manuscript should look like.
% These instructions should be used both for papers submitted for review and for final versions of accepted papers.
\end{abstract}
\section{Introduction}
\label{sec:intro}

Code-switching is a widely-occurring linguistic phenomenon in which multiple languages are used within the span of a single utterance or conversation. While large pretrained multilingual models like mBERT~\cite{devlin-etal-2019-bert} and XLM-R~\cite{conneau-etal-2020-unsupervised} have been successfully used for low-resource languages and effective zero-shot cross-lingual transfer~\cite{pires-etal-2019-multilingual,conneau-etal-2020-unsupervised,wu-dredze-2019-beto}, techniques to help these models generalize to code-switched text has not been sufficiently explored. 
%To facilitate systematic investigations into code-switched text, \citet{khanuja-etal-2020-gluecos} introduced an evaluation benchmark with a public leaderboard\footnote{\url{https://microsoft.github.io/GLUECoS/}}, \gluecos, that covers six different natural language tasks.

 %To facilitate such systematic investigations into code-switched text, \citet{khanuja-etal-2020-gluecos} introduced a new evaluation benchmark with a public leaderboard\footnote{\url{https://microsoft.github.io/GLUECoS/}}, \gluecos, that covers six different natural language tasks.

Intermediate-task training~\cite{phang2018sentence,phang-etal-2020-english} was recently proposed as an effective training strategy for transfer learning. This scheme involves fine-tuning a pretrained model on data from one or more \textit{intermediate tasks}, followed by fine-tuning on the target task. The intermediate task could differ from the target task and it could also be in a different language. This technique was shown to help with both task-based and language-based transfer; it benefited target tasks in English~\cite{vu-etal-2020-exploring} and helped improve zero-shot cross-lingual transfer~\cite{phang-etal-2020-english}.

In this work, we introduce \emph{bilingual intermediate-task training} as a reliable training strategy to improve performance on three code-switched natural language understanding tasks: Natural Language Inference (NLI), factoid-based Question Answering (QA) and Sentiment Analysis (SA). Bilingual training for a language pair \textsc{X-En} involves pretraining with an English intermediate task along with its translations in \textsc{X}. The NLI, QA and SA tasks require deeper linguistic reasoning (as opposed to sequence labeling tasks like part-of-speech tagging) and exhibit high potential for improvement via transfer learning. We present SA results for four different language pairs: Hindi-English (\HiEn), Spanish-English (\EsEn), Tamil-English (\TaEn) and Malayalam-English (\MlEn), and NLI/QA results for \HiEn.%
\footnote{These tasks present an additional challenge with the Indian languages written using transliterated/Romanized text.}
Our main findings can be summarized as follows:

%in \gluecos for code-switched Hindi-English text: Natural Language Inference (NLI) and factoid-based Question Answering (QA). Both these tasks require deeper linguistic reasoning (compared to sequence labeling tasks like part-of-speech tagging or language identification) and exhibit high potential for improvement~\cite{khanuja-etal-2020-gluecos}. These tasks present an additional challenge since Hindi is present in its Romanized form.%
%\footnote{In \citet{khanuja-etal-2020-gluecos}, only NLI and QA are evaluated on Romanized text}.
%
%Using intermediate-task training, we significantly outperform previous state-of-the-art approaches by $7.87\%$ and $20.15\%$ on mean accuracy and F1 scores for NLI and QA, respectively. Our main findings are:

\begin{asparaitem}
%\item English intermediate-task training (\eng) is beneficial for code-switched target tasks, especially for QA.
\item Bilingual intermediate-task training consistently yields significant performance improvements on NLI, QA and SA using two different pretrained multilingual models, mBERT and XLM-R. We also show the impact of translation and transliteration quality on this training scheme.
\item Pretraining using a masked language modeling (MLM) objective on real code-switched text can be used, in conjunction with bilingual training, for additional performance improvements on code-switched target tasks. We also present a code-switched MLM variant that yields larger improvements on SA compared to standard MLM. %We empirically demonstrate that MLM training is domain-sensitive and does not yield benefits when the domains of the target task and the code-switched text differ. 
%However, MLM training is domain-sensitive and does not yield performance gains when the domains of the target task and the code-switched text differ. 

%\bilingual turns out to be a more reliable training strategy to derive consistent performance improvements.
%\item On two different pretrained multilingual models, mBERT and XLM-R, we achieve superior performance using our proposed intermediate-task training schemes. 
\end{asparaitem}

\section{Methodology}
\label{sec:ittraining}

\paragraph{Intermediate-Task Training.} This scheme starts with a publicly-available multilingual model that has been pretrained on large volumes of multilingual text using MLM. This model is subsequently fine-tuned using data from one or more intermediate tasks before finally fine-tuning on code-switched data from the target tasks. 
%We elaborate on our techniques using one and multiple intermediate-tasks below.  

%In this section, we will describe our intermediate-task training techniques. Each intermediate-task training is followed by fine-tuning on the train sets of the tasks from the \gluecos benchmark.

\emph{Single Intermediate-Task Training} involves the use of existing monolingual NLI, SA and QA datasets as intermediate-tasks before fine-tuning on the respective code-switched target tasks. For a language pair \textsc{X-En}, we explored the use of three different intermediate tasks:
\begin{enumerate*}[label=(\roman*)]
    \item Task-specific data in English (\eng),
    \item Task-specific data in \textsc{X} (\xsing), and
    \item Task-specific data in both English and \textsc{X} that we call bilingual intermediate-task training (\xbilingual).
\end{enumerate*}

\emph{Multi Intermediate-Task Training} involves two intermediate-tasks (\textsc{T$_1$} and \textsc{T$_2$}) simultaneously. This training is done using two different task heads (one per task) with the pretrained models. Each batch is randomly populated with instances from tasks \textsc{T$_1$} or \textsc{T$_2$} based on the sampling technique proposed by~\cite{raffel2020transfer}. %We follow~\citet{raffel2020transfer} to sample batches from task \textsc{T$_1$} with probability $p_{\tiny{\textnormal{T}_1}} = \frac{\min(e_{\tiny{\textnormal{T}}_1}, K)}{\min(e_{\tiny{\textnormal{T}_1}}, {\tiny K}) + \min(e_{\tiny{\textnormal{T}_2}}, {\tiny K}) }$, where $e_{\tiny{\textnormal{T}_1}}$ and $e_{\tiny{\textnormal{T}_2}}$ are the number of training examples in task \textsc{T$_1$} and \textsc{T$_2$}, respectively; $p_{\tiny{\textnormal{T}_2}}$ is similarly computed. The constant $K = 2^{16}$ is used to prevent over-sampling.
We experiment with NLI and QA as the two intermediate-tasks \textsc{T$_1$} or \textsc{T$_2$} and refer to this system as \dualbilingual. We use the merged English and \textsc{Hi} datasets from \bilingual for each task. We also explored MLM training on real code-switched text as one of the tasks, in addition to the merged \textsc{X-En} task-specific intermediate-tasks (referred to as \textsc{X-En/Mlm Mutli-Task}). %(MLM finetuning was employed as an additional pretraining step in \citet{phang-etal-2020-english} and \citet{chakravarthy-etal-2020-detecting}.) 
%The next section provides details of all the datasets we used in our experiments. 

\paragraph{Code-Switched MLM.} Pretraining LMs using an MLM objective involves randomly masking tokens and reconstructing them using the surrounding context. When using MLM with code-switched text for which word-level language tags are available, we posit that masking tokens at the switching boundaries (rather than anywhere in the sentence) would explicitly make the underlying LM more sensitive to code-switching. For example:
    \centerline{\Fbox{\emph{Yeh}}$_{\texttt{\small HI}}$ \Fbox{\emph{files}}$_{\texttt{\small EN}}$ \Fbox{\emph{ko}}$_{\texttt{\small HI}}$ \Fbox{\emph{desk}}$_{\texttt{\small EN}}$ \Fbox{\emph{pe}}$_{\texttt{\small HI}}$ \Fbox{\emph{rakh}}$_{\texttt{\small HI}}$ \Fbox{\emph{do}}$_{\texttt{\small HI}}$}
    \centerline{\Fbox{\emph{Yeh}}$_{\texttt{\small HI}}$ \Fbox{\emph{files}}$_{\texttt{\small EN}}$ \Fbox{\emph{ko}}$_{\texttt{\small HI}}$ \Fbox{\emph{desk}}$_{\texttt{\small EN}}$ \Fbox{\emph{pe}}$_{\texttt{\small HI}}$ {\emph{rakh}}$_{\texttt{\small HI}}$ {\emph{do}}$_{\texttt{\small HI}}$}
    \centerline{(\textsc{Eng} translation: Put these files on the desk.)}
In the first sentence, tokens from all the boxed words can be masked (as in standard MLM), while in the second sentence (referring to code-switched MLM), only tokens from words at the boundary of a language-switch can be masked.

\begin{table*}[t!]
\begin{center}
    \scriptsize
    \setlength{\extrarowheight}{0.25pt}
    % \addtolength{\tabcolsep}{+5pt}  
    \begin{tabular}{c l|c c c|c c c|c c c|c c c}
    \toprule
        &\multirow{2}{*}{\textbf{Method}}&\multicolumn{3}{c|}{\textbf{\textsc{Es-En}} (\textsc{X: Es})}& \multicolumn{3}{c|}{\textbf{\textsc{Hi-En}} (\textsc{X: Hi})} & \multicolumn{3}{c|}{\textbf{\textsc{Ta-En}} (\textsc{X: Ta})} & \multicolumn{3}{c}{\textbf{\textsc{Ml-En}} (\textsc{X: Ml})} \\
        % &&\multicolumn{3}{c|}{\textit{Acc.}}& \multicolumn{3}{c}{\textit{F1}} \\
        \cline{3-14}
        & &  {\bf F1} & {\bf Prec.} & {\bf Rec.} &  {\bf F1} & {\bf Prec.} & {\bf Rec.} & {\bf F1} & {\bf Prec.} & {\bf Rec.} & {\bf F1} & {\bf Prec.} & {\bf Rec.} \\
        \hline
        \multirow{8}{*}{\rotatebox[origin=c]{90}{\centering \textbf{mBERT}}} & Baseline & 60.95 & 61.93 & 60.43 & 68.17 & 68.75 & 68 & 76.07 & 75.33 & 77.66 & 75.46 & 75.72 & 75.64   \\
         & \quad+\eng & 65.11 & 66.00 & 65.00 & 69.14 & 69.72 & 68.96 & 76.41 & 75.69 & 78.11 & 76.49 & 76.78 & 76.44 \\
         & \quad+\textsc{X Single-Task} & 64.69 & 65.71 & 64.57 & 68.75 & 69.37 & 68.60 & 75.78 & 74.89 & 77.80 & 75.92 & 75.96 & 76.15   \\
         & \quad+\textsc{X-En Single-Task} & 66.64 & 67.61 & 66.21 & 69.20 & 69.63 & 69.06 & 76.75 & 76.11 & 78.63 & 77.00 & 77.16 & 77.04   \\
         & \quad+\textsc{Mlm} & 62.02 & 62.93 & 61.29 & 69.89 & 70.58 & 69.76 & 76.73 & 76.14 & 78.53 & 76.13 & 76.23 & 76.24  \\
         & \quad+\textsc{Code-Switched Mlm} & 63.88 & 64.81 & 63.13 & \underline{ 70.33} & \underline{ 71.17} & \underline{ 70.10} & - & - & - & - & - & -  \\
        & \quad+\textsc{X-En/Mlm Multi-Task} & \underline{ 67.01} & \underline{{ 68.11}} & \underline{{ 66.72}} & 69.99 & 70.29 & 69.91 & \underline{{ 77.23}} & \underline{{76.6}} & \underline{{\bf 79.16}} & \underline{{77.49}} & \underline{{77.56}} & \underline{{\bf 77.58}}   \\
         %& MLM$^{+}$ + Bilingual Intermediate-Task Training & 66.17 & 65.21 & 0.96 & 79.63 & 78.27 & 1.46  \\
         \midrule
    \multirow{8}{*}{\rotatebox[origin=c]{90}{\centering \textbf{XLM-R}}} & Baseline & 66.45 & 67.45 & 65.86 & 69.37 & 69.38 & 69.46 & 75.53 & 74.56 & 77.75 & 74.14 & 74.35 & 74.15 \\
         & \quad+\eng & 67.82 & 68.89 & 67.41 & 70.23 & 70.78 & 70.09 & 76.08 & 75.41 & 77.65 &  75.14 & 75.29 & 75.42   \\
         & \quad+\textsc{X Single-Task}& 66.68 & 68.40 & 66.29 & 69.96 & 70.38 & 69.83 & 76.36 & 75.52 & 77.88 & 76.12 & 76.10 & 76.24   \\
         & \quad+\textsc{X-En Single-Task} & 68.97 & 69.79 & 68.28 & 70.23 & 70.91 & 70.01 & 76.49 & 75.90 & 77.60 & 76.68 & 76.80 & 76.62   \\
         & \quad+\textsc{Mlm} & 66.37 & 67.42 & 65.69 & 70.92 & 71.94 & 70.66 & 76.95 & 76.21 & 78.60 & 76.28 & 76.26 & 76.42  \\
         & \quad+\textsc{Code-Switched Mlm} & 67.10 & 68.30 & 66.55 & \underline{{\bf 71.74}} & \underline{{\bf 72.29}} & \underline{{\bf 71.59}} & - & - & - & - & - & - \\
         & \quad+\textsc{X-En/Mlm Multi-Task}  & \underline{{\bf 70.33}} & \underline{{\bf 71.41}} & \underline{{\bf 69.57}} & 71.08 & 71.43 & 70.97 & \underline{{\bf 77.50}} & \underline{{\bf 76.84}} & \underline{{ 78.60}} & \underline{{\bf 76.91}} & \underline{{\bf 76.94}} & \underline{{76.98}}   \\
         %& MLM$^{+}$ + Bilingual Intermediate-Task Training & 65.77 & 64.69 & 0.92 & 83.58 & 81.55  & 1.27  \\
        %  \midrule
         %\multicolumn{14}{c}{\textbf{Comparison with Previous Work} (max scores)}\\
         \hline
         &\textbf{Our Best Models} (Max) & 71.7 & 72.8 & 71.3 & 72.6 & 73.2 & 72.4 & 78 & 77 & 79 &78 & 78 & 78\\
         %&$\rho$-gCM~\cite{pratapa-etal-2018-word} & 64.6 &- &- & - & -& -& -&- & -& - & - & -\\
         %&Stock XLM-R~\cite{srinivasan-2020-msr} &-&- &- & 72.6 & -& -& -&- & -& - & - & -\\
         %&TweetEval~\cite{gupta-etal-2021-task} & - & - & - & - & - & - & 77 & 76 & 79 & 78 & 78 & 78\\
         \bottomrule
         
    \end{tabular}
    % \addtolength{\tabcolsep}{-5pt}
\end{center}

\caption{Main results for Sentiment Analysis. Best results for each model are underlined and the overall best results are in bold. All scores are weighted averages and are further averaged over five runs with random seeds.}%Since \gluecos tasks are low-resource, we report the mean and standard deviation over 5 runs with different seeds along with the maximum score on the test sets. 

\label{tab:sentiment}
\end{table*}

\section{Experimental Setup}
\label{sec:setup}

\paragraph{Datasets.} The \HiEn NLI dataset is from a recent code-switched benchmark \gluecos~\cite{khanuja-etal-2020-new} comprising 1.8K/447 training/test examples, respectively. The \HiEn factoid-based QA dataset~\cite{chandu-etal-2018-code} is also from \gluecos, consisting of 259/54 training/test question-answer pairs (along with corresponding context), respectively. While code-switched NLI and QA tasks were only available in \HiEn, we show SA results for four language pairs. The \EsEn SA dataset~\cite{vilares-etal-2016-en} in \gluecos consists of 2.1K/211/211 examples in train/dev/test sets, respectively. The \HiEn SA dataset~\cite{patwa-etal-2020-semeval}
%
%\footnote{We chose this dataset over \gluecos as it is more data-rich. However, the \EsEn dataset could not be used as we did not have access to the test labels for evaluation.}
%
comprises 15K/1.5K/3K code-switched tweets in train/dev/test sets, respectively. The train/dev/test sets in the \TaEn SA dataset~\cite{chakravarthi-etal-2020-corpus} and \MlEn SA dataset~\cite{chakravarthi-etal-2020-sentiment} comprise 9.6K/1K/2.7K and 3.9K/436/1.1K code-switched YouTube comments, respectively. As the evaluation metric, we use accuracies for NLI and SA over two (entailment/contradiction) and three labels (positive/negative/neutral), respectively,  and F1 scores for the QA task. 
 
As intermediate tasks for NLI and QA, we used \textsc{En} and \textsc{Hi} versions of the MultiNLI dataset~\cite{williams-etal-2018-broad} (with 250/10K examples in the train/dev sets) and the SQuAD dataset~\cite{rajpurkar-etal-2016-squad} (consisting of 82K/5K question-answer pairs in its train/dev sets), respectively. The \textsc{Hi} translations for SQuAD (in Devanagari) are available in the \textsc{xtreme}~\cite{Hu2020XTREME} benchmark. We used \emph{indic-trans}~\cite{Bhat2014fire} to transliterate the \textsc{Hi} translations, since NLI and QA in \gluecos use Romanized \textsc{Hi} text. For \EsEn and \HiEn SA, we used TweetEval~\cite{barbieri-etal-2020-tweeteval} dataset (63K sentences in total) and its translations in \textsc{Es} and \textsc{Hi} generated via MarianMT\footnote{Implementation used: \url{http://bit.ly/MarianMT}} \cite{junczys-dowmunt-etal-2018-marian-fast} and IndicTrans MT~\cite{ramesh2021samanantar} respectively. %This dataset was chosen because of its domain similarity with the code-switched tweets. However, the quality of data dropped significantly when translated into \textsc{Ta} and \textsc{Ma}.
For \TaEn and \MlEn, we used the positive, negative and neutral labelled sentences from the SST dataset~\cite{socher-etal-2013-recursive} (100K instances) as the intermediate task. The \textsc{Ta} and \textsc{Ml} translations were also generated using the IndicTrans MT system. The translations were further transliterated using \citet{Bhat2014fire} for \textsc{Hi} and the Bing Translator API\footnote{\url{http://bit.ly/azureTranslate}} for \textsc{Ta} and \textsc{Ml}. 
%\textsc{Ma} in the Romanized script, we transliterate the translated data using \cite{Bhat2014fire} (for \textsc{Hi}) and the Bing Translator API\footnote{\url{http://bit.ly/azureTranslate}} (for \textsc{Ta} and \textsc{Ma})\footnote{Alternatives like \textit{indic-trans} and Google Translate API either left residual characters or generated accented characters during transliteration. Please refer to Appendix X\todo{add} for examples}. 
For MLM pretraining, we accumulate text appearing in various prior work resulting in a total of roughly 185K, 66K, 139K and 40K real code-switched sentences in \HiEn, \EsEn, \TaEn and \MlEn, respectively. Please refer to Appendix A for more details. (All these datasets will be publicly released upon publication.)  

\paragraph{Model Details.} mBERT is a transformer model~\cite{vaswani2017attention} pretrained using MLM on the Wikipedia corpus of 104 languages. XLM-R uses a similar training objective as mBERT but is trained on orders of magnitude more data from the CommonCrawl corpus spanning 100 languages and yields competitive results on low-resource languages\cite{conneau-etal-2020-unsupervised}. %It gives better results on cross-lingual tasks including tasks on low-resource languages~\cite{conneau-etal-2020-unsupervised}.
We use the \emph{bert-base-multilingual-cased} and \emph{xlm-roberta-base} models from the Transformers library~\cite{wolf2019huggingface}. We refer readers to Appendix~\ref{ap:tuning} for more implementation details. %and the respective heads for various tasks. 

% To further verify this, we down-sample the English MultiNLI training data to have only as many examples as \gluecos NLI data. We finetune mBERT using this smaller corpus. Poor validation accuracy is observed, even though mBERT has seen lots of English in its pretraining stage.

%varying MLM size and mono/bi lingual main task

\section{Results and Discussion}
\label{sec:result}

%Table~\ref{tab:sentiment} provides a comprehensive evaluation of our proposed techniques. We show that our techniques significantly outperform the baselines and prior work across tasks (NLI and QA) and multilingual models (mBERT and XLM-R\footnote{As in \citet{chakravarthy-etal-2020-detecting}, we also find that XLM-R baseline on \gluecos NLI does not converge. However, our techniques mitigate this artefact.}). Since the \gluecos test sets are small in size, we report mean accuracies and F1 scores over 5 runs with different seeds for NLI and QA, respectively. 
%(along with max scores and standard deviations). %\footnote{We do not provide an MLM-based baseline as it significantly hurts QA performance. Exact numbers are in Appendix~\ref{ap:mlm}.} 

\paragraph{SA Results.} 
Table~\ref{tab:sentiment} shows our main results for SA on \textsc{Es-En}, \textsc{Hi-En}, \textsc{Ta-En} and \textsc{Ml-En}. We observe that \xbilingual outperforms both \eng and \xsing with up to $9.33\%$ relative improvements over the baseline for \textsc{Es}. For all language pairs (other than \HiEn), \textsc{X-En/Mlm Multi-Task} is the best-performing system. 
While MLM training consistently improves over the baseline, we see that code-switched MLM provides additional performance gains compared to standard MLM training for \textsc{Es-En} and \textsc{Hi-En}. We do not report code-switched MLM results for \textsc{Ta} and \textsc{Ma} since we do not have access to language labels or a trained language identification system for either language. %(Refer to Appendix Z\todo{fill} for more detailed MLM experiments.)%
%\footnote{In \textsc{Es-En}, roughly half of the code-switched sentences described in section~\ref{sec:setup} do not exhibit switching. For a fair comparison of MLM and Code-Switched MLM, these sentences are discarded resulting in 30K sentences overall. For MLM numbers using the entire data, refer to Appendix Z\todo{Fill}.} 
%An analysis of the MLM data showed that $45\%$ of all tokens belonged to words on a switching boundary, and thus the masking probability was increased from $0.15$ to $0.3$ to ensure the same level of masking.
%
%Apart from a few exceptions, all our techniques yield improvements over the baseline, with \textsc{X-En Single-Task} systems outperforming the \textsc{En Single-Task} and \textsc{X Single-Task} counterparts. In the case of Spanish-English and Malayalam-English \textsc{X-En Single-Task} outperforms the only MLM systems, and for the other language pairs the performance is quite close. For all language pairs apart from Hindi-English, \textsc{X-En/Mlm Multi-Task} gives the best performance and the same technique can scale well to other tasks such as NLI and QA.

%Table~\ref{tab:sentiment} shows that our best models match the performance of if not outperform the models used in previous work. 
 A notable advantage of our bilingual training is that we outperform (or match) previous state-of-the-art with an order of magnitude less data. Our best \EsEn system yields an F1 of $71.6$ compared to  \citet{pratapa-etal-2018-word} with an F1 of $64.6$.
 %\footnote{We do not use the results by \citet{khanuja-etal-2020-gluecos} in their paper as it did not match those on the leaderboard.}. 
 For \HiEn, our best F1 ($72.6$) matches the $2^\text{nd}$-ranked system~\cite{srinivasan-2020-msr} on SentiMix 2020~\cite{patwa-etal-2020-semeval}. For \TaEn and \MlEn, our best systems match the score of the best TweetEval model in~\citet{gupta-etal-2021-task}. While prior work required roughly 17M sentences in \EsEn, 2.09M sentences in \HiEn and 60M tweets to train TweetEval for \textsc{Ta} and \textsc{Ma}, we use 192K, 180K, 330K and 240K sentences for the four respective languages.  
 %with gain in \textsc{Ta} F1 and precision from $(77, 76)$ to $(78, 77)$.
%
%

%The advantage of our techniques is that our techniques require much lesser data. In the case of Spanish\footnote{We do not use the results by \citet{khanuja-etal-2020-gluecos} in their paper as it did not match those on the leaderboard.}, \citet{pratapa-etal-2018-word} use about 17M sentences, whereas our best system uses only 192K sentences (English tweets, their Spanish translations, and the MLM data). Similarly, in Hindi, \citet{srinivasan-2020-msr} use 2.09M code-switched sentences in contrast to our best system that uses 180K sentences. This brings us to share the $2^{nd}$ position on the Hindi SentiMix 2020 task~\cite{patwa-etal-2020-semeval}. In the case of Tamil and Malayalam, TweetEval approach used by \citet{gupta-etal-2021-task} uses 60M tweets in MLM pretraining whereas our approach uses 330K and 240K sentences for Tamil and Malayalam respectively.

%Our techniques are much more generalizable to a variety of tasks and language pairs, as shown by the Sentiment Analysis experiments. Even in highly resource constrained settings like Code Switched Malayalam, our models perform at par with other models. This might not be possible with setups like $\rho$-gCM \citet{pratapa-etal-2018-word} which can end up needing up plenty of data to generate reliable embeddings.

%
\begin{table}[t!]
\begin{center}
    \scriptsize
    \setlength{\extrarowheight}{0.35pt}
    \addtolength{\tabcolsep}{-2.5pt}  
    \begin{tabular}{c l|c c|c c}
    \toprule
        &\multirow{2}{*}{\textbf{Method}}&\multicolumn{2}{c|}{\textbf{\textsc{Gluecos} NLI} }& \multicolumn{2}{c}{\textbf{\textsc{Gluecos} QA} } \\
        % \textit{(F1)}
        % &&\multicolumn{3}{c|}{\textit{Acc.}}& \multicolumn{3}{c}{\textit{F1}} \\
        \cline{3-6}
        & &  {\bf Max} & {\bf Mean}  &  {\bf Max} & {\bf Mean}  \\
        \hline
        \multirow{6}{*}{\rotatebox[origin=c]{90}{\centering \textbf{mBERT}}} & Baseline & 61.07 & 57.51  & 66.89 & 64.25  \\
        & \quad+\textsc{Mlm} & 59.94 & 58.75 & 60.8 & 58.28    \\
         & \quad+\eng & 62.40 & 60.73 & 77.62 & 75.77   \\
         & \quad+\hindi & 63.73 & 62.09  & 79.63 & 76.77   \\
         & \quad+\bilingual & 65.55 & 64.10 & 81.61 & 79.97   \\
         & \quad+\dualbilingual & \dunderline{0.5pt}{\bf 66.74} & {65.3}  & \dunderline{0.5pt}{83.03} & \dunderline{0.5pt}{80.25} \\
         & \quad+\mlmbilingual & 66.66 & \dunderline{0.5pt}{\bf 65.61}& 81.05 & 79.11 \\
         %& MLM$^{+}$ + Bilingual Intermediate-Task Training & 66.17 & 65.21 & 0.96 & 79.63 & 78.27 & 1.46  \\
         \midrule
    \multirow{6}{*}{\rotatebox[origin=c]{90}{\centering \textbf{XLM-R}}} & Baseline & - & - & 56.86 & 53.22 \\
        & \quad+\textsc{Mlm} & - & - & 45.9 & 42.34   \\
         & \quad+\eng & {66.22} & 63.91  & 82.04 & 80.92  \\
         & \quad+\hindi & 63.24 & 61.73& 81.48 & 80.55   \\
         & \quad+\bilingual & 65.01 & {64.37}  & 82.41 & 81.36  \\
         & \quad+\dualbilingual & 64.49 & 64.35  & \dunderline{0.5pt}{\bf 83.95} & \dunderline{0.5pt}{\bf 82.38}  \\
         & \quad+\mlmbilingual & \dunderline{0.5pt}{66.66} & \dunderline{0.5pt}{65.01}  & 82.1 & 80.44  \\
         %& MLM$^{+}$ + Bilingual Intermediate-Task Training & 65.77 & 64.69 & 0.92 & 83.58 & 81.55  & 1.27  \\
         \midrule
         \multicolumn{6}{c}{\bf Previous work on \textsc{Gluecos}}\\
         \hline
         \multicolumn{2}{l|}{mBERT \cite{khanuja-etal-2020-gluecos}\tablefootnote{Dataset changes led to outdated results in the paper. We report latest numbers after consulting the authors of \gluecos.}} & 59.28 & 57.74  & 63.58 & 62.23   \\
         \multicolumn{2}{l|}{mod-mBERT \cite{chakravarthy-etal-2020-detecting}} & 62.41 & - & - & -   \\
        %  \multicolumn{2}{l|}{Our Best Models} & & & & & &  \\
        % $^{\small\dagger}$
         \bottomrule
         
    \end{tabular}
    \addtolength{\tabcolsep}{+2.5pt}
\end{center}
\caption{Main results for NLI and QA for \HiEn.}% Best results for each model are underlined and the overall best results are in bold. Since \gluecos tasks are low-resource, we report the mean and standard deviation over 5 runs with different seeds along with the maximum score on the test sets. 
\label{tab:main}
\end{table}
\paragraph{NLI/QA Results.} Table~\ref{tab:main} shows our main results for the NLI and QA tasks in \HiEn. Among the \textsc{Single-Task} systems, \bilingual performs the best (based on mean scores) on both NLI\footnote{Like \citet{chakravarthy-etal-2020-detecting}, we also find that XLM-R baseline/\textsc{Mlm} on \gluecos NLI does not converge.} and QA. Using a merged \HiEn dataset for \bilingual training was critical. (Sequentially training on English followed by \textsc{Hi} resulted in poor performance, as shown in Appendix~\ref{ap:seq}.) 
%Another interesting observation is that XLM-R benefits more from \eng while mBERT benefits more from \hindi, compared to the baseline. This could be attributed to XLM-R having encountered Romanized \textsc{Hi} text during its pretraining unlike mBERT.  
The \textsc{Multi-Task} systems yield additional gains. Using both NLI and QA as intermediate tasks benefits both NLI and QA for mBERT and QA for XLM-R, and corroborates observations in prior work~\cite{tarunesh2021meta,phang-etal-2020-english}. 
%\citet{tarunesh2021meta} observed that NLI and QA mutually benefit from each other in a multi-task framework using mBERT, while \citet{phang-etal-2020-english} showed that QA benefits from NLI with XLM-R in a multi-task framework, but the converse does not hold. Our experiments corroborate both these observations. 
Although intermediate-task training is beneficial across tasks, the relative improvements in QA are higher than that for NLI (see Appendix~\ref{ap:egs} for some QA examples). We conjecture this is due to varying dataset similarity between intermediate-tasks and target tasks~\cite{vu-etal-2020-exploring}. %In QA, this similarity is higher and in NLI the conversational nature and large premise lengths reduces this similarity (see Appendix~\ref{ap:datasets} for examples). Our observation that task and domain similarity play a crucial role in transferability is consistent with \citet{vu-etal-2020-exploring}. 
The effect of domain similarity is more pronounced with MLM training resulting in variations between absolute $1.5$-$2\%$ (details in Appendix~\ref{ap:mlm}).

\paragraph{Translation and Transliteration Quality.} 
%The NLI and QA intermediate tasks are obtained from \textsc{xtreme} that uses an in-house machine translation system. 
To assess the impact of both translation and transliteration quality on NLI and QA performance, we use two small datasets XNLI~\cite{conneau-etal-2018-xnli} and XQuAD~\cite{artetxe-etal-2020-cross} for which we have manual \textsc{Hi} (Devanagari) translations. In Table~\ref{tab:trans-translit}, we compare the performance of \bilingual using manual translations with translations from the Google Translate API\footnote{\url{https://cloud.google.com/translate}}, and also transliterations from this API with those from \emph{indic-trans}. %Table~\ref{tab:trans-translit} shows the performance of \bilingual for different choices of translation and transliteration outputs. 
(Appendix~\ref{ap:trans} contains additional details, experiments, and a similar study for SA.)
% \iffalse
% Our \textsc{Hi} data for the intermediate tasks was obtained from \textsc{xtreme}; they used an in-house machine translation system to translate the English datasets to \textsc{Hi} (in Devanagari) and we further transliterated it. To assess the impact of both translation and transliteration quality on final performance, we use two small datasets for NLI and QA from XNLI~\cite{conneau-etal-2018-xnli} and XQuAD~\cite{artetxe-etal-2020-cross} for which we have manual \textsc{Hi} (Devanagari) translations. The NLI/QA datasets contain 4.2K/2.3K training instances, respectively. We used the corresponding English text to generate \textsc{Hi} translations using the Google Translate API.\footnote{\url{https://cloud.google.com/translate}} For XNLI, the premises and hypotheses were directly translated and for XQuAD we adopted the same translation procedure listed in~\citet{Hu2020XTREME}. We use two transliteration tools, \emph{indic-trans} and the transliteration output that is a by-product of the Google Translate API. Table~\ref{tab:trans-biling} shows the performance of \bilingual for different choices of translation and transliteration outputs. (Appendix~\ref{ap:trans} shows a similar comparative analysis using \hindi.)
% \fi
We observe that manual translation and transliteration using the Google API performs best. For NLI, manual translation followed by transliteration using \emph{indic-trans} outperforms machine-translation and transliteration by the Google API, while for QA the trend is reversed. This indicates that translation and transliteration quality have varying impacts depending on the task. Interestingly, Tables~\ref{tab:main} and~\ref{tab:trans-translit} show that the performance after training on small amounts of bilingual manually translated data is statistically comparable to the \hindi systems that use much larger amounts of text. 
\begin{table}[t!]
    \scriptsize
    \centering
    \setlength{\extrarowheight}{0.35pt}
    % \addtolength{\tabcolsep}{-0.75pt}  
    \begin{tabular}{l|c c c}
    \toprule
         \bf Translate $-$ Transliterate & \bf Max & \bf Mean & \bf Std. \\
    \hline
        & \multicolumn{3}{c}{\textsc{Gluecos} NLI  \textit{(acc.)}}\\
        \cline{2-4}
         Manual $-$ Google Translate API & 62.24 & 61.6 & 0.62 \\
         Manual $-$ \emph{indic-trans} & 62.09 & 59.71 & 1.37 \\
         Google Translate API (both) & 60.18 & 58.59 & 1.07 \\
         \hline
         & \multicolumn{3}{c}{\textsc{Gluecos} QA \textit{(F1)}}\\
         \cline{2-4}
         Manual $-$ Google Translate API & 79.32 & 77.33 & 2.22 \\
         Manual $-$ \emph{indic-trans} & 78.09 & 76.35 & 1.36 \\
         Google Translate API (both) & 78.44 & 76.72 & 1.22 \\
         \bottomrule
    \end{tabular}
    % \addtolength{\tabcolsep}{+0.75pt}  
    \caption{Effect of translation and transliteration quality on intermediate-task training for NLI and QA.}
\label{tab:trans-translit}
\end{table}

\section{Related Work}

\citet{pires-etal-2019-multilingual} and \citet{hsu-etal-2019-zero} showed that mBERT is effective for \HiEn part-of-speech tagging and a reading comprehension task on synthetic code-switched data, respectively. This was extended for a variety of code-switched tasks by~\citet{khanuja-etal-2020-gluecos}, where they showed improvements on several tasks using MLM pretraining on real and synthetic code-switched text. \citet{chakravarthy-etal-2020-detecting} further improved the NLI performance of mod-mBERT by including large amounts of in-domain code-switched text during MLM pretraining. %They also explored the use of XLM-R by data-augmentation with SNLI~~\cite{bowman-etal-2015-large} and XNLI, but failed to generate improvements on the test set.
Intermediate task-training has proven to be effective for many NLP target tasks~\cite{pruksachatkun-etal-2020-intermediate, vu-etal-2020-exploring}, as well as cross-lingual zero-shot transfer from English tasks on multilingual models such as XLM-R~\cite{phang-etal-2020-english} and mBERT~\cite{tarunesh2021meta}. \citet{gururangan-etal-2020-dont} empirically demonstrate that pretraining is most beneficial when the domains of the intermediate and target tasks are similar, which we observe as well. Differing from their recommendation of domain adaptive pretraining using MLM on large quantities of real code-switched data, we find intermediate-task training using significantly smaller amounts of labeled data to be more consistently beneficial across tasks and languages. In contrast to very recent work~ \cite{gupta-etal-2021-task} that reports results using a Roberta-based model trained exclusively for SA and pretrained on 60M English tweets, we present a bilingual training technique that is consistently effective across tasks and languages while requiring significantly smaller amounts of data.   

%The influenced of the domain of MLM data on performance is consistent with \citet{gururangan-etal-2020-dont}. Further, MLM on real code-switched data can be viewed as a type of domain adaptive pretraining. However, code-switched tasks are so low resource that pretraining on unlabeled task data is not useful. Our training approaches involving intermediate-task training using labelled data are more effective and consistent across tasks and language pairs. \citet{gupta-etal-2021-task} present results for task-specific training by using Roberta-based model pretrained on 60M english tweet and extensively fine-tuned on sentiment analysis. In contrast, our study is more comprehensive, comprising of monolingual and bilingual, single and multi-task setups. Our training paradigms are consistently effective across tasks and language pairs while using substantially lesser data points without compromising the performance.

\section{Conclusion}
This is the first work to demonstrate the effectiveness of intermediate-task training for code-switched NLI, QA and SA on different language pairs, and present code-switched MLM that consistently benefits SA more than standard MLM. Given the promising results, for future work, we plan to continue exploring pretraining strategies using selective masking and task-adaptive techniques.

%We substantially improve over previous state-of-the-art with up to $8\%$ and $20\%$ absolute improvements on NLI and QA, respectively. For future work, we plan to extend our investigation to include more language pairs and more intermediate tasks.

\clearpage
% \bibliography{anthology,custom}
% \bibliographystyle{acl_natbib}

\clearpage
\appendix
\section*{Appendix}

\section{Additional Dataset and Model Details}
\label{ap:datasets}
\subsection{\textsc{xtreme} Translate-Train Datasets}
    As mentioned previously, we use the MultiNLI and SQuAD v1.1 data from the translate-train sets of the \textsc{xtreme} benchmark\footnote{MultiNLI: \url{https://storage.cloud.google.com/xtreme_translations/XNLI/translate-train/en-hi-translated.tsv}}\textsuperscript{,}\footnote{SQuAD: \url{https://storage.cloud.google.com/xtreme_translations/SQuAD/translate-train/squad.translate.train.en-hi.json}.}. The Romanized version of these datasets are generated using the \emph{indic-trans} tool~\cite{Bhat2014fire} starting from their Devanagari counterparts. For NLI, we directly transliterated the premise and hypothesis. For QA, the context, question and answer were transliterated and the answer span was corrected. This was done by calculating the start and stop indices of the span, and then, doing a piece-wise transliteration. We finally checked if  the context-span matched the answer text. All instances passed this check. To benefit future work in this direction, we provide the transliterated datasets\footnote{Available at: \url{http://bit.ly/finalCSdata}}. In Tables~\ref{tab:data-nli} and~\ref{tab:data-qa}, we present some examples from our datasets. 

\subsection{Masked Language Modelling}
We use a corpus of 64K real code-switched sentences by pooling data listed in prior work~\cite{singh-etal-2018-twitter,swami2018corpus,chandu-etal-2018-language}; we will call it \generalcs. We supplant this text corpus with an additional 28K code-switched sentences mined from movie scripts (referred to as \moviecs), which is more similar in domain to \gluecos NLI. We further used code-switched text from \citet{patwa-etal-2020-semeval}, \citet{bhat-etal-2017-joining}, and \citet{patro-etal-2017-english} resulting in a total of 185K \HiEn sentences. For \EsEn 66K real code-switched sentences were pooled from prior work \cite{patwa-etal-2020-semeval, solorio-etal-2014-overview, alghamdi-etal-2016-part, ws-2018-approaches, vilares-etal-2016-en}. We pooled roughly 130K real code-switched sentences in \textsc{Ta} \cite{chakravarthi-etal-2020-corpus, chakravarthi-etal-2021-findings-shared-task,banerjee-etal-2018-dataset, hasoc2020}, and 40K code-switched sentences in \textsc{Ml} \cite{chakravarthi-etal-2020-sentiment,chakravarthi-etal-2021-findings-shared-task, hasoc2020}.

In some of the MLM experiments, we used code-switched data from the \gluecos datasets. In NLI, we split various dialogues from premises (based on the \#\# separator) and discarded the ones with less than 5 words (due to insufficient context for MLM). This gives us around 6.5K code-switched sentences. Similarly, we split the sentences from the passage in the QA task. This gives us an additional 4.1K code-switched sentences. 
\iffalse 
We create a general corpus of code switched sentences (referred to as \generalcs), by mixing code switched data from~\citet{singh-etal-2018-twitter}, \citet{swami2018corpus} and~\citet{chandu-etal-2018-language}. This gives rise to a corpus of 64,882 CS sentences. Ishan's CS corpus (referred to as \moviecs), obtained from movie scripts shares a similar domain as the the \gluecos NLI corpus. It has a total of 28,080 CS instances. From this, we create a held-out dev set of 1,500 instances.

We combine \moviecs with the \generalcs and use 3,000 sentences as held-out dev set. We also combine \moviecs with the dialogues (separated by \textit{\#\#}) in the premises of the \gluecos NLI training data. All dialogues with less than 5 words are removed, since they tend to lack context for meaningful MLM. From the remaining, 2,000 examples are moved to the held-out dev set.

The \generalcs is also mixed with the code switched contexts from the \gluecos training set. Each sentence from a corpus (obtained by splitting the text on `.') is treated as a separate training example. This leads to a total of 31034 CS instances in the combined corpus.
\fi
\subsection{Number of Model Parameters}
 The mBERT model comprises 179M parameters with the MLM head comprising 712K parameters. The XLM-R model comprises 270M parameters with an MLM head with 842k parameters. For both models, the NLI (sequence classification) and QA heads comprise 1536 parameters each. For SA (sequence classification) the head comprises of 2304 parameters.
\section{Hyperparameter Tuning}
\label{ap:tuning}
In all experiments, we have used the AdamW algorithm~ \cite{loshchilov2019decoupled} and a linear scheduler with warm up for the learning rate. These experiments were run on a single NVIDIA GeForce GTX 1080 Ti GPU. Some crucial fixed hyperparameters are: \texttt{\small learning\_rate = 5e-5}, \texttt{\small adam\_epsilon = 1e-8}, \texttt{\small max\_gradient\_norm = 1}, and \texttt{\small gradient\_accumulation\_steps = 10}.

\subsection{Intermediate-Task Training}
The training for all the main intermediate-task experiments was carried out for 4 epochs and the model with the highest performance metric on the task dev set was considered (all the metrics stagnated after a certain point in training). For NLI + QA tasks, two separate models were stored depending on the performance metric on the respective dev set. No hyperparameter search was conducted at this stage. During bilingual training, the batches were interspersed---equal number of examples from English and Romanized \textsc{Hi} within each batch. In the single-task systems, we used \texttt{\small batch\_size = 8} and \texttt{\small max\_sequence\_length = 128} for NLI, \texttt{\small batch\_size = 8} and \texttt{\small{max\_sequence\_length = 256}} for SA, \texttt{\small batch\_size = 4} and \texttt{\small max\_sequence\_length = 512} for QA. During multi-task training, the \texttt{\small max\_sequence\_length} was set to the maximum of the aforementioned numbers and the respective batch-sizes. Any multi-task training technique requires at least 14-15 hours for validation accuracy to stagnate. Single task intermediate training requires 4-5 hours for monolingual versions and 8-9 hours for the bilingual version.  SA data being smaller in size requires 8-9 hours for multitask, 4-5 hours for bilingual intermediate task and 1-2 hours for monolingual intermediate task. The \texttt{\small{logging\_steps}} are set to  approximately 10\% of the total steps in an epoch.

\subsection{Fine-tuning on \gluecos NLI \& QA Tasks}
The base fine-tuning files have been taken from the \gluecos repository\footnote{\url{https://github.com/microsoft/GLUECoS}}. Given that there no dev sets in \gluecos, and that the tasks are low-resource, we use train accuracy in NLI and train loss in QA as an indication to stop fine-tuning. Manual search is performed over a range of epochs to obtain the best test performance. For NLI, we stopped fine-tuning when training accuracy is in the range of 70-80\% (which meant fine-tuning for 1-4 epochs depending upon the model and technique used). For QA, we stopped when training loss reached $\sim$ 0.1. Thus, we explored  3-5 epochs for mBERT and 4-8 epochs for XLM-R. We present the statistics over the best results on 5 different seeds. We used \texttt{\small batch\_size = 8} and \texttt{\small max\_sequence\_length = 256} for \gluecos NLI\footnote{The sequence length was doubled as compared to the intermediate-task training to incorporate the long premise length of \gluecos NLI. This resulted in higher accuracy.} and \texttt{\small batch\_size = 4} and \texttt{\small max\_sequence\_length = 512} for \gluecos QA. All our fine-tuning runs on \gluecos take an average of 1 minute per epoch.

\subsection{Fine-tuning on downstream SA tasks}
The dev set, being available for all language pairs was used to find the checkpoint with best F1 score, and this model was used for evaluation on the test set. The mean values were presented after carrying out the above procedure for 6 different seeds. The \texttt{\small{logging\_steps}} are set to  approximately 10\% of the total steps in an epoch. Each epoch takes around 1 minute for \textsc{Ta}, \textsc{Ma} and \textsc{Es}, 2 minutes for \textsc{Hi} (SemEval). 

\section{Experiments on Translation and Transliteration Quality }
\label{ap:trans}
We combined the test and dev sets of XNLI to get the data for intermediate-task training. We discarded all examples labelled \emph{neutral} and instances where the crowdsourced annotations did not match the designated labels\footnote{This was achieved via the \emph{match} Boolean attribute~\cite{conneau-etal-2018-xnli}}. After this, we were left with roughly 4.2K/0.5K instances in the train/dev sets, respectively (the dev set is used for early stopping during intermediate-task training). For XNLI, the premises and hypotheses were directly translated and for XQuAD we adopted the same translation procedure listed in~\citet{Hu2020XTREME}.
\begin{table}[t!]
    \centering
    \scriptsize
    \setlength{\extrarowheight}{1pt}
    % \addtolength{\tabcolsep}{-0.75pt}  
    \begin{tabular}{l|c c c}
    \toprule
         \bf Translate $-$ Transliterate & \bf Max & \bf Mean & \bf Std. \\
    \hline
        & \multicolumn{3}{c}{\gluecos NLI \textit{(acc.)}}\\
        \cline{2-4}
         Manual $-$ Google Translate API & 61.05 & 59.63 & 0.96 \\
         Manual $-$ \emph{indic-trans} & 59.50 & 59.26 & 0.23 \\
         Google Translate API (both) & 60.12 & 58.54 & 1.53 \\
         \hline
         & \multicolumn{3}{c}{\gluecos QA \textit{(F1)}}\\
         \cline{2-4}
         Manual $-$ Google Translate API & 73.50 & 71.36 & 1.46 \\
         Manual $-$ \emph{indic-trans} & 70.19 & 68.26 & 1.26 \\
         Google Translate API (both) & 72.73 & 69.63 & 2.2 \\
         \bottomrule
    \end{tabular}
    % \addtolength{\tabcolsep}{+0.75pt}  
    \caption{Effect of translation and transliteration quality during intermediate-task training on \gluecos results.}
    \label{tab:trans-hi}
\end{table}
\begin{table}[t!]
    \centering
    \scriptsize
    \setlength{\extrarowheight}{1pt}
    \addtolength{\tabcolsep}{-1pt}  
    \begin{tabular}{c|c|c|c c c}
    \toprule
         \bf Task &\bf Model & \bf Translit Tool & \bf F1 & \bf Prec. & \bf Rec. \\
    \hline
        \multirow{4}{*}{\textsc{Ta Single-Task}} & \multirow{2}{*}{mBERT} & \emph{indic-trans} & 75.42&	74.72&	76.62  \\
        & & Bing API & 75.78&	74.89&	77.8 \\
        \cline{2-6}
        &  \multirow{2}{*}{XLM-R} & \emph{indic-trans} & 75.51&	74.87&	76.66  \\
        & & Bing API & 76.36&	75.52&	77.88 \\
        \midrule
        
        \multirow{4}{*}{\textsc{Ml Single-Task}} & \multirow{2}{*}{mBERT} & \emph{indic-trans} & 74.7	&74.82&	74.71  \\
        & & Bing API &75.92 &	75.96&	76.15\\
        \cline{2-6}
        &  \multirow{2}{*}{XLM-R} & \emph{indic-trans} & 74.68&	74.82&	74.66  \\
        & & Bing API & 76.12&	76.1&	76.24 \\
        
         \bottomrule
    \end{tabular}
    \addtolength{\tabcolsep}{+1pt}  
    \caption{Effect of transliteration quality of intermediate-tasks on SA results. Scores are weighted averages further averaged over 5 random runs.}
    \label{tab:trans-tm-ml}
\end{table}

\begin{figure*}[t!]
    \centering
    \includegraphics[width=12cm]{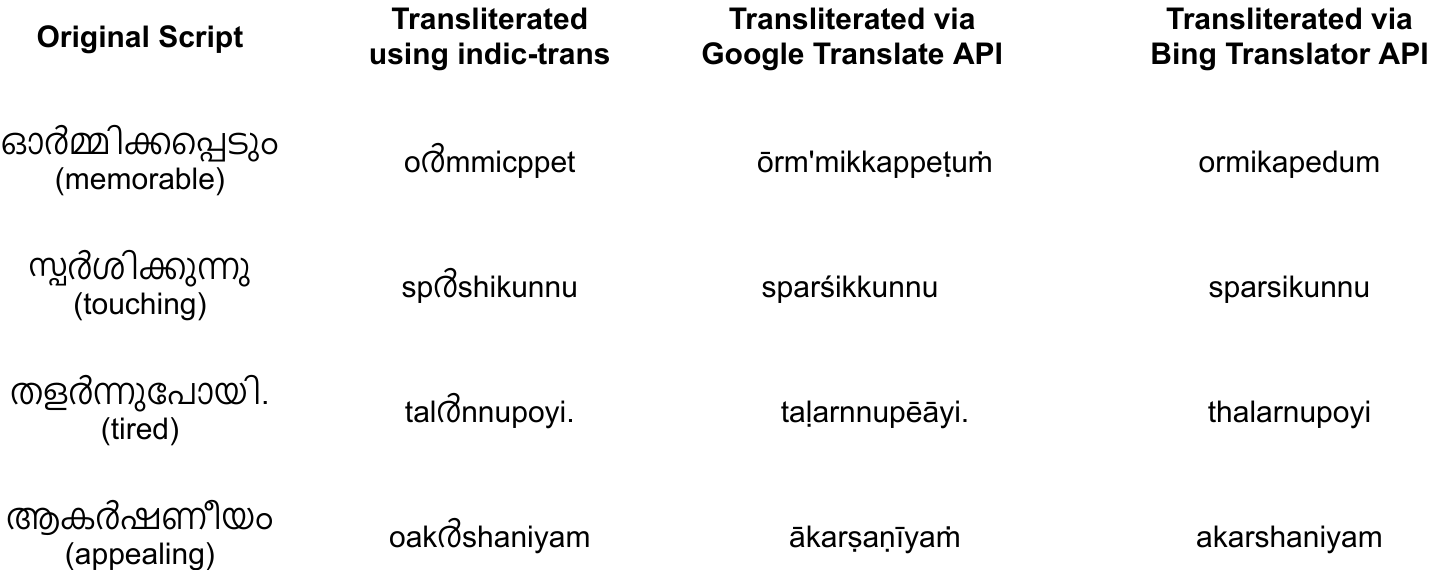}
    \caption{Different transliterations for some descriptive words in \textsc{Ma}; \textit{indic-trans} leaves some residual characters in native script and Google translate API generates accented characters.}
    \label{fig:translit}
\end{figure*}

Similar to Table~\ref{tab:trans-translit} for bilingual text, we present a similar analysis of romanized \textsc{Hi} text in Table~\ref{tab:trans-hi} for mBERT. All the relative trends remain the same. Table~\ref{tab:trans-tm-ml} shows the impact of transliteration on SA. Figure~\ref{fig:translit} illustrates different \textsc{Ma} transliterations.

In Tables~\ref{tab:data-nli} and \ref{tab:data-qa}, we show some instances from the datasets. The color-coded transliterations indicate that \emph{indic-trans} often uses existing English words as transliterations. While for some specific (uncommon) words that is helpful, in most cases it leads to ambiguity in the sentence meaning (shown in blue). Further, these ambiguous words (in blue) are far more common in the \textsc{Hi} language, and thus, have a greater impact on model performance. We also note that transliteration of these common words in the \gluecos dataset matches closely with the transliteration produced using the Google Translate API. Further, there is not a lot of difference between the machine and human translations, which might be due to translation bias. %However, \citet{Hu2020XTREME} use an in-house machine translation system which may differ in translation quality.

\begin{table}[t!]
    \centering
    \scriptsize
    \setlength{\extrarowheight}{1pt}
    % \addtolength{\tabcolsep}{-0.75pt}  
    \begin{tabular}{l|c c c}
    \toprule
         \bf Intermediate-Task Paradigm & \bf Max & \bf Mean & \bf Std. \\
    \hline
        & \multicolumn{3}{c}{\gluecos NLI \textit{(acc.)}}\\
        \cline{2-4}
         \eng & 62.40 & 60.73 & 1.78 \\
         \hindi &63.73 & 62.09 & 0.99  \\
         \bilingual & 65.55 & 64.1 & {0.89} \\
         Sequential Training: \textsc{En} $\rightarrow$ \textsc{Hi} & 62.02 & 59.94 & 1.83 \\
         \hline
         & \multicolumn{3}{c}{\gluecos QA \textit{(F1)}}\\
         \cline{2-4}
         \eng & 77.62 & 75.77 & 1.79  \\
         \hindi & 79.63 & 76.77 & 1.86 \\
         \bilingual & 81.61 & 79.97 & {1.29}  \\
         Sequential Training: \textsc{En} $\rightarrow$ \textsc{Hi}  & 76.23 & 73.69 & 1.78 \\
         \bottomrule
    \end{tabular}
    % \addtolength{\tabcolsep}{+0.75pt}  
    \caption{Sequential bilingual training yields poorest performance on both the tasks using the mBERT model.}
    \label{tab:seq}
\end{table}

\begin{table}[t]
    \centering
    \scriptsize
    \setlength{\extrarowheight}{1.0pt}
    % \addtolength{\tabcolsep}{-2.65pt}  
    \begin{tabular}{c|c|c c c}
    \toprule
         \bf Language & \bf MLM Data & \bf Max & \bf Mean & \bf Std. \\
    \hline
        & & \multicolumn{3}{c}{\gluecos NLI \textit{(acc.)}}\\
        \cline{3-5}
        - & \generalcs & 59.94 & 58.75& 0.93\\
        \hline
        \multirow{2}{*}{\textsc{En}}   & - &62.40 & 60.73 & 1.78\\
         & \generalcs & \dunderline{0.5pt}{65.07} & \dunderline{0.5pt}{62.84} & 1.93\\
         \hline
         \multirow{5}{*}{\textsc{Hi-En}}   & - & 65.55 & 64.1 & {0.89}\\
         & \generalcs & 65.22 & 64.19 & 1.22 \\
        %  & \moviecs & 60.1 & 57.3 & 2.19 \\
             & \textsc{General + Movie cs} & \dunderline{0.5pt}{\bf 66.67} & \dunderline{0.5pt}{\bf 65.61} & 0.86\\
             \multirow{2}{*}{}& \textsc{General + Movie cs} &
              \multirow{2}{*}{66.17} & \multirow{2}{*}{65.21} &\multirow{2}{*}{0.96}\\
             & \textsc{+} \gluecos \textsc{nli cs}  & & & \\
         \midrule
         && \multicolumn{3}{c}{\gluecos QA \textit{(F1)}}\\
         \cline{3-5}
          - & \generalcs & 59.26 & 57.84& 1.29\\
        \hline
         \multirow{2}{*}{\textsc{En}}   & - & \dunderline{0.5pt}{77.62} & \dunderline{0.5pt}{75.77} & 1.79  \\
         & \generalcs & 76.23 & 75.49 & 1.03 \\
         \hline
         \multirow{5}{*}{\textsc{Hi-En}}   & - & 81.61 & 79.97 & {1.29}\\
         & \generalcs & \dunderline{0.5pt}{\bf 83.03} & \dunderline{0.5pt}{\bf 80.38} & 1.68\\
         & \textsc{General + Movie cs} & 81.05 & 79.11 & 1.40
         \\
         \multirow{2}{*}{}& \textsc{General + Movie cs} & \multirow{2}{*}{79.63} & \multirow{2}{*}{78.27} &\multirow{2}{*}{1.46}\\
             & \textsc{+} \gluecos \textsc{qa cs}  & & & \\
          
         \bottomrule
    \end{tabular}
    % \addtolength{\tabcolsep}{+2.65pt}  
    \caption{Performance on different variations of MLM + intermediate-task training of mBERT. We underline the relatively best model and bold-face the model with the highest performance for each task.}
    \label{tab:mlm}
\end{table}

\section{Alternate Bilingual Training Paradigm}
\label{ap:seq}

\begin{table*}[t!]
    \centering
    \scriptsize
    \setlength{\extrarowheight}{1.0pt}
    \begin{tabularx}{\textwidth}{c C c c}
    \toprule
            \textbf{Language} & \textbf{Premise/ Hypothesis} & \textbf{Label} & \textbf{Dataset} \\
            \hline
         \textsc{Hi-En} & {\bf \textsc{premise:}} \textsc{Clerk} : Yeh ? \#\# \textsc{Clerk} : Yeh toh guzar gaya . Haadsa ho gaya ek . \#\# \textsc{Devi} : Iski ye file hai humaare paas . Kuch paise baaki hain , ek do books bhi hain uski jo lautani hain . Aap pata de sakte hain ? \#\# \textsc{Clerk} : Number hai ghar ka . Ye addresss hai unka Allahabad mein . Note karo . Par bolna mat kahin ki maine diya hai . \#\#  \newline {\bf \textsc{hypothesis:}} \textsc{Clerk} ki kuch files \textsc{Devi} ke paas hain.  & contradictory & \gluecos NLI\\
         \hline
         \textsc{En} & {\bf \textsc{premise:}} Split Ends a Cosmetology Shop is a nice example of appositional elegance combined with euphemism in the appositive and the low key or off-beat opening. \newline {\bf \textsc{hypothesis:}} Split Ends is an ice cream shop. &	entailment & MultiNLI/ XNLI \\
         \hline
         \textsc{Hi}  (Google$^\diamond$) & {\bf \textsc{premise:}}  split ends ek \addmarked{kosmetolojee} shop epositiv \marked{aur} kam kunjee ya oph-beet \addmarked{opaning} mein vyanjana ke saath sanyukt eplaid laality ka ek achchha udaaharan hai. \newline	{\bf \textsc{hypothesis:}} split \marked{ends} ek \addmarked{aaisakreem} shop \marked{hai}.  & entailment & Translation$^{\dagger}$\\
         \hline
         \textsc{Hi}  (Google$^\diamond$) & {\bf \textsc{premise:}} split inds ek \addmarked{kosmetolojee} shop samaanaadhikaran shishtata \marked{aur} kam kunjee ya of-beet \addmarked{opaning} mein preyokti ke mishran ka ek achchha udaaharan \marked{hai}. \newline {\bf \textsc{hypothesis:}} split \marked{ends} ek \addmarked{aaisakreem} kee dukaan \marked{hai}.
        & entailment & XNLI\\
         \hline
         \textsc{Hi}  (indic$^\star$) & {\bf \textsc{premise:}} split inds ek \addmarked{cosmetology} shop samaanaadhikaran shishtataa \marked{or} kam kunjee yaa of-beet opening main preyokti ke mishran kaa ek acha udhaaharan \marked{he}.\newline {\bf \textsc{hypothesis:}} split \marked{ands} ek \addmarked{icecream} kii dukaan \marked{he}. & entailment & XNLI\\
         \bottomrule
         
         \hline
    \end{tabularx}
    \caption{NLI examples from some of our datasets. $^{\dagger}$: obtained by translation of the second row using Google Translate API. $^\diamond$: transliterated using Google Translate API, $^\star$: transliterated using \emph{indic-trans}~\cite{Bhat2014fire}. In \marked{blue}, we show some of the words with ambiguous transliteration by \emph{indic-trans} and their counterparts. In \addmarked{purple}, we show some words that are better transliterated by \emph{indic-trans}. Best viewed in color.}
    \label{tab:data-nli}
\end{table*}

To further probe the effectiveness of bilingual task-specific intermediate training, we train on the monolingual task data sequentially. That is, we first train on the English corpus and then on the Romanized \textsc{Hi} corpus from \textsc{xtreme}. We observe that this results in relatively poor performance on both tasks, even worse than the monolingual counterparts. This indicates that just additional data is not sufficient and our scheme of simultaneous bilingual training is important to achieve good performance. Table~\ref{tab:seq} shows the performance of various techniques and validates our observations.

\section{MLM + Intermediate-Task Training}
\label{ap:mlm}

Table~\ref{tab:mlm} provides a comprehensive summary of our experiments exploring intermediate-task training of mBERT in conjunction with MLM in a multi-task framework. As described in Section~\ref{sec:setup}, our setup is similar to \citet{raffel2020transfer}. The first key take-away from Table~\ref{tab:mlm} is that intermediate training using MLM on code-switched data alone is not effective (first row of each task).

NLI benefits from MLM in a multi-task setup in both monolingual and bilingual settings. Further, we note that adding in-domain \moviecs data yields additional improvements. We do not present the experiment with MLM on only the \moviecs data that is relatively smaller in size because of its inferior performance. This shows that sufficient amount of in-domain data is needed for performance gains, and augmenting out-of-domain with in-domain code-switched text can be effective.

In the case of QA, MLM does not improve performance in the monolingual setting, although the mean scores are statistically close. In the bilingual setting, we see a clear improvement using \generalcs for MLM training. However, using both \generalcs and \moviecs for MLM results in significant degradation of performance. We believe that this is because of the fact that the domain of the passages in \gluecos QA is similar to the \HiEn blog data present in \generalcs. However, the \moviecs dataset comes from a significantly different domain and thus hurts performance. This indicates that in addition to the amount of unlabelled real code-switched text, when using MLM training, the domain of the text is very influential in determining the performance on \gluecos tasks.

\begin{table*}[t!]
    \centering
    \scriptsize
    \setlength{\extrarowheight}{1.0pt}
    \begin{tabularx}{\textwidth}{c C c c}
    \toprule
            \textbf{Language} & { \textbf{QA Context}} &  \textbf{Dataset} \\
            \hline
         \textsc{Hi-En} & Mitashi ne ek Android Tv ko Launch kiya hain. Jise tahat yeh Tv Android Operating System par chalta hain. Iski Keemat Rs. 51,990 rakhi gayi hain. Ab aaya Android TV Mitashi Company ne Android KitKat OS par chalne wale Smart TV ko Launch kar diya hain. Company ne is T.V. ko 51,990 Rupees ke price par launch kiya hain. Agar features ki baat kare to is Android TV ki Screen 50 inch ki hain, Jo 1280 X 1080 p ka resolution deti hain. USB plug play function ke saath yeh T.V. 27 Vidoe formats ko support karta hain. Vidoe input ke liye HDMI Cable, PC, Wi-Fi aur Ethernet Connectivity di gyi hain. Behtar processing ke liye dual core processor ke saath 512 MB ki RAM lagayi gyi hain. Yeh Android TV banane wali company Mitashi isse pahle khilaune banane ka kaam karti thi. Iske alawa is company ne education se jude products banane shuru kiye. 1998 mein stapith huyi is company ne Android T.V. ke saath-saath India ki pahli Android Gaming Device ko bhi launch kiya hain.  & \gluecos QA\\
         \hline
         \textsc{En} &Their local rivals, Polonia Warsaw, have significantly fewer supporters, yet they managed to win Ekstraklasa Championship in 2000. They also won the country’s championship in 1946, and won the cup twice as well. Polonia's home venue is located at Konwiktorska Street, a ten-minute walk north from the Old Town. Polonia was relegated from the country's top flight in 2013 because of their disastrous financial situation. They are now playing in the 4th league (5th tier in Poland) -the bottom professional league in the National – Polish Football Association structure. & SQuAD/XQuAD \\
         \hline
         \textsc{Hi}  (Google$^\diamond$) & unake sthaaneey pratidvandviyon, \addmarked{poloniya  voraso} ke paas kaaphee kam samarthak hain, phir bhee ve 2000 \marked{mein} ekastraklaasa \addmarked{chaimpiyanaship} jeetane \marked{mein} kaamayaab rahe. unhonne 1946 \marked{mein} desh kee \addmarked{chaimpiyanaship} bhee jeetee, \marked{aur} do baar \marked{kap} bhee jeeta. \addmarked{poloniya}  ka ghareloo sthal konaveektarsaka street par sthit \marked{hai}, jo old taun se uttar \marked{mein} das \addmarked{minat} kee paidal dooree par \marked{hai}. apanee vinaashakaaree vitteey sthiti ke kaaran \addmarked{poloniya}  ko 2013 \marked{mein} desh kee sheersh udaan se hata diya gaya tha. ab ve neshanal (polish polish esosieshan)  sanrachana \marked{mein} 4 ven leeg (polaind \marked{mein} 5 ven star) \marked{mein} khel rahe hain. &  Translation$^{\dagger}$\\
         \hline
         \textsc{Hi}  (Google$^\diamond$) & unake sthaaneey pratidvandviyon, \addmarked{poloniya  vaaraso}, ke paas kaaphee kam samarthak hain, phir bhee ve 2000 \marked{mein} ekalastralaasa \addmarked{chaimpiyanaship} jeetane \marked{mein} kaamayaab rahe. unhonne 1946 \marked{mein} raashtriy \addmarked{chaimpiyanaship} bhee jeetee, \marked{aur} saath hee do baar \marked{kap} jeete. \addmarked{poloniya}  ka ghar konaveektarsaka street par sthit \marked{hai}, jo old taun se uttar \marked{mein} das \addmarked{minat} kee paidal dooree par \marked{hai}. \addmarked{poloniya}  ko 2013 \marked{mein} unakee kharaab vitteey sthiti kee vajah se desh kee sheersh udaan se hata diya gaya tha. ve ab botam profeshanal leeg ke 4th leeg (polaind \marked{mein} 5 ven star) neshanal polish futabol esosieshan sanrachana \marked{mein} khel rahe hain. & XQuAD\\
         \hline
         \textsc{Hi}  (indic$^\star$) & unke sthaneey pratidwandviyon, \addmarked{polonia  warsaw}, ke paas kaaphi kam samarthak hai, phir bhi ve 2000 \marked{main} ecrestlasa \addmarked{championships} jeetne \marked{main} kaamyaab rahe. unhone 1946 \marked{main} rashtri \addmarked{championships} bhi jiti, \marked{or} saath hi do baar \marked{cap} jite. \addmarked{polonia}  kaa ghar konwictarska street par sthit \marked{he}, jo old toun se uttar \marked{main} das \addmarked{minute} kii paidal duuri par \marked{he}. \addmarked{polonia}  ko 2013 \marked{main} unki karaab vittiya sthiti kii vajah se desh kii sheersh udaan se hataa diya gaya tha. ve ab bottm profeshnal lig ke 4th lig (poland \marked{main} 5 wein str) neshnal polish footbaal association sanrachana \marked{main} khel rahe hai. & XQuAD\\
         \bottomrule
         
         \hline
    \end{tabularx}
    \caption{QA examples from some of our datasets. $^{\dagger}$: obtained by translation of the second row using Google Translate API. $^\diamond$: transliterated using Google Translate API, $^\star$: transliterated using \emph{indic-trans}~\cite{Bhat2014fire}. In \marked{blue}, we show some of the words words with ambiguous transliteration by \emph{indic-trans} and their counterparts. In \addmarked{purple}, we show some words that are better transliterated by \emph{indic-trans}. Best viewed in color. }
    \label{tab:data-qa}
\end{table*}

\begin{table*}[t!]
    \centering
    \scriptsize
    \setlength{\extrarowheight}{1.0pt}
    \begin{tabularx}{\textwidth}{c C c c}
    \toprule
            \textbf{Language} & \textbf{Sentence} & \textbf{Label} & \textbf{Dataset} \\
            \hline
    \textsc{En} & It's definitely Christmas season! My social media news feeds have been all about Hatchimals since midnight! Good luck parents! & positive & TweetEval \\
    \textsc{Hi}$^\diamond$ & yeah nishchit roop se christmas ka mausam hai! mera social media news feed adhi raat se hatchimal ke baare mein hai! mata-pita ko shubhkamnayen! & positive & Translation$^{\dagger}$\\
    \textsc{Es} & ¡Es definitivamente la temporada de Navidad! Mis noticias en las redes sociales han sido todo acerca de Hatchimals desde medianoche! ¡Buena suerte padres! & positive & Translation$^{\ddagger}$ \\
    \textsc{Ml}$^\diamond$ & ith theerchayayum chrismas seesonnan, ente social media news feads ardharathri muthal hachimalsine kurichan! &  positive & Translation$^{\dagger}$\\ \textsc{Ta}$^\diamond$ & itu nichchayam christumus column! nalliravu muthal enathu samook utaka seithi oottngal anaithum hatchimals patriadhu! petrors nalvazthukal! & positive & Translation$^{\dagger}$\\
     \textsc{En} & the story and the friendship proceeds in such a way that you 're watching a soap opera rather than a chronicle of the ups and downs that accompany lifelong friendships . & negative & SST \\
    \textsc{Hi}$^\diamond$ & kahani or dosti is tarah se aage badhati hai ki op jeevan bhar ki dosti ke saath aane vale utaar-chadhav k kram k bajay ek dharavahik dekh rahe hain & negative & Translation$^{\dagger}$\\
     \textsc{Es}  & la historia y la amistad proceden de tal manera que estás viendo una telenovela en lugar de una crónica de los altibajos que acompañan a las amistades de toda la vida. & negative & Translation$^{\ddagger}$\\
    \textsc{Ml}$^\diamond$ & ajeevanantha sauhradangalil undakunna uyarchayeyum thazhchayeyum kurichulla oru kathayalla, marich oru sopp opera kanunna reethiyilan kathayum souhrdavum munnot pokunnath. & negative & Translation$^{\dagger}$\\
    \textsc{Ta}$^\diamond$ & kadhaiyum natpum vaazhnaal muzhuvathum natputan inaintha etra erakangalin kalavarisai cottlum neengal oru soap oberov paarkkum vagaiyil selgiradhu. & negative & Translation$^{\dagger}$\\
         \bottomrule
         
         \hline
    \end{tabularx}
    \caption{Sentiment analysis examples from our datasets. $^{\dagger}$: obtained by translation of the corresponding \textsc{En} sentence using IndicTrans MT~\cite{ramesh2021samanantar}. $^\ddagger$: obtained by translation of the corresponding \textsc{En} sentence using MarianMT.  $^\diamond$: transliterated using Bing Translator API.}
    \label{tab:data-sa}
\end{table*}

For both these tasks, we observe a common trend: Adding code-switched data from the training set of \gluecos tasks degrades performance. This could be due to the quality of training data in the \gluecos tasks. Each dialogue in the NLI data does not have a lot of content and is highly conversational in nature. In addition to this, the dataset is also very noisy. For example, a word \emph{`humko'} is split into characters as \emph{`h u m k o'}. Thus, MLM on such data may not be very effective and could hurt performance. The reasoning in the case of QA is different: For a significant portion of the train set, the passage is obtained using \emph{DrQA - Document Retriever module}\footnote{\url{https://github.com/facebookresearch/DrQA}}~\cite{chen-etal-2017-reading}. These passages are monolingual in nature and thus not useful for MLM training with code-switched text.
\section{\textsc{Es-En MLM} Results}
The \EsEn MLM dataset has a lot of sentences with very little or no switching. The language identity detection system hence ends up marking such sentences, as either having only English or only \textsc{Es} words. Thus, these utterances become unfit for Code-Switched MLM, and thus are discarded. For a fair comparison to vanilla MLM, we presented the numbers with only these sentences. Upon using the complete En-Es MLM corpus, we obtained an F1 of 62.57 using mBERT and 67.6 using XLM-R on the sentiment analysis test set of \gluecos En-Es.
% \clearpage

% \section{NLI + QA + MLM Joint Training}
% One natural extension to our technique is joint training with NLI, QA and MLM on code-switched data as intermediate-tasks in our multi-task framework. However, we have not explored this training for two reasons---\begin{enumerate*}[label=(\roman*)]
% \item as highlighted in Appendix~\ref{ap:mlm}, NLI and QA tasks in \gluecos benefit from different MLM (from different domains) data and data from other domains can hurt performance, and
% \item  the MLM heads have orders of magnitude more parameters and given our current computational capacity, joint training on all the three tasks was infeasible.
% \end{enumerate*} 
\section{Example Outputs from \gluecos QA}
\label{ap:egs}
The examples below are outputs on the \gluecos QA test set. Our techniques improve the ability of the multilingual models to do deeper reasoning. Consider the following question which is correctly answered by the baseline mBERT:

\noindent \textsc{q:} \textit{``Continuum ye feature kaunsi company ne launch kiya hai?''} (``Which company launched the continuum feature?'')

\noindent \textsc{a:} \textit{Microsoft}

%\noindent It is answered correctly by mBERT finetuned on \gluecos QA test set. 
However the question,

\noindent\textsc{q:} \textit{``Microsoft ke kaunse employee nein Continuum ka kamaal dikhaya?''} ( ``Which employee of Microsoft showed the wonders of Continuum?'')

\noindent \textsc{a:} \textit{Joe Belfear}

\noindent is not correctly answered. Our \eng system (trained on English SQuAD) answers both questions correctly. \\

\noindent Our \eng system is also able to correctly answer questions that invoke numerical answers, such as: \\

\noindent \textsc{q:} \textit{``4G SIM paane ke kitne option hai??''} (``How many options are there to get a 4G SIM?'').

\noindent \textsc{a:} 2

%mBERT without any intermediate training answers the above question incorrectly. mBERT with intermediate training on English SQuAD lists the correct answer in the top-3 predictions. mBERT with intermediate training on bilingual SQuAD is able to answer the question perfectly.
\end{document}